# Chapter 1
# AI Competitions and Benchmarks: The life cycle of challenges and benchmarks

Gustavo Stolovitzky, Julio Saez-Rodriguez, Julie Bletz, Jacob Albrecht, Gaia Andreoletti, James C. Costello, Paul Boutros

> ... it is known with certainty that there is scarcely anything which more greatly excites noble and ingenious spirits to labors which lead to the increase of knowledge than to propose difficult and at the same time useful problems through the solution of which, as by no other means, they may attain to fame and build for themselves eternal monuments among posterity.
>
> *Johann Bernoulli, June 1696* [1]

**Abstract** Data Science research is undergoing a revolution fueled by the transformative power of technology, the Internet, and an ever-increasing computational capacity. The rate at which sophisticated algorithms can be developed is unprecedented, yet they remain outpaced by the massive amounts of data that are increasingly available to researchers. Here we argue for the need to creatively leverage the scientific research and algorithm development community as an axis of robust innovation. Engaging these communities in the scientific discovery enterprise by critical assessments, community experiments, and/or crowdsourcing will multiply oppor-

DREAM Challenges e-mail: `gustavo.stolo@gmail.com`

Heidelberg University, Faculty of Medicine, and Heidelberg University Hospital, Institute for Computational Biomedicine, Bioquant, Heidelberg, Germany e-mail: `julio.saez@uni-heidelberg.de`

Sage Bionetworks, Seattle, WA, USA e-mail: `julie.bletz@sagebase.org`

Sage Bionetworks, Seattle, WA, USA e-mail: `jake.albrecht@sagebase.org`

Department of Pediatrics, University of California, San Francisco, San Francisco, CA, USA e-mail: `Gaia.Andreoletti@ucsf.edu`

Department of Pharmacology, University of Colorado Anschutz Medical Campus, Aurora, CO, USA e-mail: `james.costello@cuanschutz.edu`

Department of Urology and Human Genetics, University of California, Los Angeles, CA, USA e-mail: `pboutros@mednet.ucla.edu`

tunities to develop new data-driven, reproducible and well-benchmarked algorithmic solutions to fundamental and applied problems of current interest. Coordinated community engagement in the analysis of highly complex and massive data has emerged as one approach to find robust methodologies that best address these challenges. When community engagement is done in the form of competitions — also known as challenges — the validation of the analytical methodology is inherently addressed, establishing performance benchmarks. Finally, challenges foster open innovation across multiple disciplines to create communities that collaborate directly or indirectly to address significant scientific gaps. Together, participants can solve important problems as varied as health research, climate change, and social equity. Ultimately, challenges can catalyze and accelerate the synthesis of complex data into knowledge or actionable information, and should be viewed a powerful tool to make lasting social and research contributions.

## 1.1 Brief history of crowdsourcing

The idea of leveraging a community of experts and non-experts to solve a scientific problem has been around for hundreds of years. One early example is the 1714 British Board of Longitude Prize, which was to be awarded to the person who could solve arguably the most important technological problem of the time: to determine a ship's longitude at sea [2]. After eluding many established scientists of the time, the prize was awarded to John Harrison for his invention of the marine chronometer. There are two important take home messages from the Longitude Prize example. One is the fact that the winner of the prize was John Harrison, an unknown carpenter and clock-maker, and not a more recognized scientist of that era. The second key idea is that the problem was posed as an open participation challenge, what we refer to today as crowdsourcing. When it comes to analyzing large or novel datasets, it is very likely that the analytical methods and breakthroughs that get the most useful signal may reside with groups other than the data generators or the most established and best published groups in a field.

Coined by Jeff Howe in an article in Wired Magazine [3], crowdsourcing mixes the bottom-up creative intelligence of the community that volunteers solutions with the top-down management of the organization that poses the problem. Crowdsourcing has been used in many contexts such as business (the design of consumer products) [4], journalism (the collection of information), and peer-review (in the evaluation of patent applications). Here, we are interested in the application of crowdsourcing to computational problems in science and technology.

## 1.2 Types of crowdsourcing

Different types of crowdsourcing exist. Generally speaking, crowdsourcing can refer to efforts in which the crowd provides or generates data (e.g. patients provide their medical information) to be mined by others, or alternatively, to initiatives in which the crowd actively works on solving a problem from established data [5].

With respect to active crowdsourcing, there are several types. There is labor-focused crowdsourcing, where the job is made open and any individual willing to work can take up the job [4]. A well-known example of labor-focused crowdsourcing is the 'mechanical turk' service run by Amazon and named after an 18th century chess automaton with a hidden human inside. The mechanical turk approach provides an online workforce that allows people to complete work, or "Human Intelligence Tasks", in exchange for a small amount of money [6].

A problem can also be crowdsourced in a manner that divides it up into a set of separate smaller tasks to solve. Crowdsourcing of data annotation and curation in Bioinformatics can be handled well with this approach. This scheme has also been applied to provide pathway resources [7, 8], reconstruct the human metabolic network [9], annotate molecular interactions in Mycobacterium tuberculosis [10], and identify critical errors in ontologies [11].

In contrast to labor-focused forms of crowdsourcing, where people are paid for their efforts, there are forms of crowdsourcing where individuals volunteer their effort because of their interest in the project or cause. An example of this is the crowdsourced approach taken to the development of Wikipedia, the Internet's largest and most popular general reference source. In some instances, such as Wikipedia and the protein structure game Foldit [12], participants contribute their time and intellectual capacity, while in other examples, such as the Folding@home [13] and rosetta@home [14] projects, participants provide computational power from their personal equipment to help solve the problem. Interestingly, Foldit evolved from rosetta@home, when participants realized that, in some cases, they could intuitively see better structures than those predicted computationally. NASA has also leveraged the power of crowdsourcing with its Citizen Science Projects [15] in which NASA scientists and interested members of the public (the citizen scientists) work together in dozens of astronomy problems such as finding new features in Jupiter's atmosphere and searching for exoplanets, to name a few.

In some instances, crowdsourcing can be implemented in the form of a game [16] in order to maximize the number of solvers who work on the problem and increase the likelihood that they will stay engaged. For example, in the Foldit project, the problem of determining protein structure is transformed into an entertaining game. Such gamification, where one applies game-design elements to allow an enjoyable experience, has proven a spectacular approach to raise participant numbers and interest. It also leads to impact: Foldit's 57,000 players provided useful results that matched or outperformed algorithmically computed solutions [12]. Foldit was followed by a similarly popular project, EteRNA [17], where more than 26,000 participants provided an RNA sequence that fits in a given shape. The best designs, as chosen by the community, were then tested experimentally [18, 12]. More recently, the EteRNA platform was used to task the community of RNA enthusiasts to design RNA molecular sensors [19]. The thousands of designed RNA molecules were synthesized and experimentally tested using high-throughput biochemical assays as part of the "game", in an iterative process that closed the loop between the real world and the online world. A diversity of tasks can be embedded in existing games. For example, a mini-game in the EVE Online game engaged 322,006 gamers to provide 33 million classifications. Combining these with deep learning led to largely improve image-classification [20]. In another example, the Borderlands Science mini-game in Borderlands 3 [21] engaged over a million participants to solve 50 million tasks within three months. The results of the mini-game are aggregated to improve a reference alignment of millions of ribosomal sequences for gut flora.

Crowdsourcing projects are also effective for collecting new ideas or directions that may be needed to solve a tough problem. These are referred to as "ideation" challenges, and the Board of Longitude Prize mentioned above falls within this category. More recently the Longitude Prize 2014 [22] built on the success of its predecessor to address the problem of antibiotic resistance through the creation of point-of-care test kits for bacterial infections. Amongst many other ideation prizes, the XPRIZE Qualcomm Tricoder prize [23] encouraged participants to develop a handheld wireless device that monitors and diagnoses health conditions.

When combined with crowdsourcing, benchmarking activities become a powerful approach to rapidly develop solutions that exceed the state of the art. At the most basic level, a performance benchmark requires a task, a metric, and a means of authenticating the result. In this modality of crowdsourcing, data is provided to participants along with the particular question to be addressed. Often, the organizers of such challenges will have "ground truth" or "Gold Standard" data that is known only to them and allows them to objectively score the methods that participants develop. Participants submit their solutions so that the organizers evaluate against the Gold Standard data. In this way, it is possible to find the best available method to solve the problem posed, and the participants can get an objective assessment of their methods. The organization of these efforts requires clear scoring metrics for evaluation of the solutions and availability of non-public datasets for use as Gold Standards. With the addition of a time constraint, participants are motivated to work concurrently, thus increasing the opportunities for collaboration and idea exchange.

Perhaps the first benchmarking challenge, posed to the many, but for which the solution is known only to a few, was used by Johann Bernoulli in the 17th century [24] to compare methods to solve a mathematical problem. Johann Bernoulli had solved what today is known as the problem of the brachistochrone, consisting of determining the shape of the non vertical curve that would make the time taken by a bead sliding on it under the effect of gravity, reach the other extreme in the shortest time. This challenge was advertised in an article [1] published in June 1696 using the words cited in the epigraph of this chapter. The incentive was then, as it is now, to make or break reputations, as indicated by Bernoulli when he states that through the solution of these challenges, solvers *"may [...] build for themselves eternal monuments among posterity."*. In the end, five solutions were submitted, including an anonymous one bearing an English postmark which was likely Newton's submis-sion. So, who won this challenge? In some ways it was a tie, as all 5 solutions were correct. But Jakob Bernoulli's solution distinguishes itself as using methods that would eventually pave the way to the development of the calculus of variations.

Benchmarking challenges have played a key role in the evolution of many areas of AI and predictive modeling. The standard ingredients of benchmarking competitions include a stated problem, enrolled competitors, a public dataset on which competitors train their models, and a scoring methodology which assesses the accuracy of the predictions on a private, hidden test set. This methodology is known in some quarters as the Common Task Framework [25, 26]). In 1986 DARPA adopted the the Common Task Framework as a new paradigm to advance machine translation research. Over time, incremental improvements have accumulated to yield much of the technology that is taken for granted today, such as optical character recognition, instant automatic translation, dictation, and commanding computers by voice. The combination of this objective approach to benchmarking with predictive modeling culture has been described as the "secret sauce" of machine learning [26].

The field of "protein folding" provides a sterling example of how benchmarking can nucleate communities and be a powerful driving force that lead to the solution of fundamental problems in science. Proteins are large, complex molecules that are essential to all life forms. Proteins are polymers composed of a linear chain of subunits called aminoacids which interact with each other and with the surrounding water. These interactions make the polymer fold in 3D space until it reaches a stable minimum energy configuration. The function of proteins depend to a large degree on the 3D configuration it adopts. Figuring out the 3D shape of proteins is known as the "protein folding problem", a grand challenge in biology whose solution defied scientists for the past 50 years. CASP (Critical Assessment of protein Structure Prediction), a benchmarking competition to assess the accuracy of protein structure prediction methods, has played a pivotal role in advancing the field [27]. Since its first edition in 1994, CASP has provided a biennial forum to benchmark the performance of structure prediction methods by comparing the predicted fold of proteins against their actual 3D structures known only by the organizers. In this way CASP can establish the state of the art in this field of science. Recently, CASP was at the center of a major scientific advance, by enabling the demonstration that the AI system called AlphaFold achieved an unprecedented accuracy in solving the protein structures proposed by the CASP organizers [28]. This breakthrough demonstrates the impact that benchmarking competitions can have in motivating the community and taking the pulse of the progress towards the solution of fundamental scientific problems.

The previous examples aim to illustrate how benchmarking challenges can 1) encourage the community to focus their efforts to solve important problems in science and technology by providing enticing incentives for participation, 2) provide a fair and objective comparison of the submitted solutions and help set the state of the art of a field, and 3) accelerate the pace of research by crowdsourcing a problem to a community of experts. These types of challenges, in which a competition is a means to a collaborative effort, are sometimes called collaborative competitions, critical assessments, or community experiments, and are the focus of this Chapter.

## 1.3 Collaborative competitions

A collaborative competition - often called a challenge or **coopetition** - is a specific form of crowdsourcing that has grown in popularity amongst research scientists during the past few decades. This kind of challenge leverages the use of leaderboards that allow participants to monitor their performance ranking with respect to others. Leaderboards also provide real-time feedback throughout a challenge to assess how their performance changes as their model evolves. Collaborative competitions also offer incentives, such as monetary prizes and/or the opportunity to co-author scientific publications that result from the challenge. General factors that lead to successful challenges include low entry barriers, continuous stimulation from the organizers, small intermediate milestones/rewards, an important problem with high economical and/or societal impact, and opportunities to get recognition or potentially a job offer.

It has been observed in many contexts such as those described above that there is a wisdom in the crowd [29], a sort of emerging intelligence in which an aggregation of solutions proposed by teams, as long as they are working relatively independently, is often more accurate than any of the single solutions. This community wisdom gives real meaning to the notion of collaboration by competition.

Many collaborative competitions, covered in Chapters 7, 8, and 9, have emerged over the last few decades. These cover multiple areas of science and technology, ranging from fundamental to applied questions in machine learning (Kaggle [30], Codalab [31]), structural biology such as protein structure prediction (CASP [27] – Critical Assessment of protein Structure Prediction, CAPRI33), to asses algorithms developed to interpret human genetic variation (CAGI [32] - the Critical Assessment of Genome Interpretation), to the assessment of data coming from a specific experimental technology (e.g. BioCreative for natural language processing [33] or MICCAI [34] for Medical Image Analysis). Collaborative competitions also provide a framework to evaluate software pipelines to process different data types, such as RGASP (RNA-seq Genome Annotation Assessment Project) that runs a competition to evaluate software to align partial transcript reads to a reference genome sequence, a key step in RNA-seq data processing [35]. Other initiatives started with a narrow focus and then broadened their spectrum. For example the DREAM Challenges originally addressed the inference of gene regulatory networks from experimental data [36], and hence the name DREAM: Dialogue for Reverse Engineering Assessment and Methods. However, over the years DREAM has evolved to address challenges ranging from regulatory genomics to translational medicine [37]. Recently, NeurIPS, the premiere conference in machine learning, has started challenges (NeurIPS Competitions) on a broad range of topics. These initiatives are often driven by academic efforts, although companies or other institutions and disease foundations (e.g. Prize4Life, Alzheimer's Research UK and the Arthritis Foundation) also run or support them.

## 1.4 The ingredients for a collaborative competition

Central to any crowdsourced challenge is the scientific question that the challenge intends to address. Not all questions are amenable to the challenge framework and not all questions have the potential to spark a community to action. The question must be fundamental to a field of research. Often times challenges are incentivized with academic publications, cash prizes and travel grants, as well as the interest of researchers to help to advance their field. Good challenge questions are conceptually straightforward and attract researchers from many fields of study who apply their specific principles and methods to address the question. For example, in the DREAM Network Inference Challenge [38], the task was to infer and benchmark transcription factor-to-target gene relationships. Because networks and graphic modeling are fundamental to many areas, researchers from Computational Biology, Engineering, Computer Science and Physics, to name a few, applied methods developed in their fields to address this challenge. Another example includes the Higgs Boson Machine Learning Challenge [1], which was timely and attracted the largest number of Kaggle competitors at that time. Participants were eager to learn how physics experiments to discover new particles were conducted and were motivated to contribute. All top ten ranking participants were not particle physicists and novel methods from oceanography and other fields were contributed. This trend, namely the convergence of experts in different disciplines to try to solve a timely problem, has been observed in other challenges. Inspiring new approaches to a fundamental question is central to crowdsourced challenges, thus relating science and technology questions to principles shared across many disciplines helps drive innovation.

From a practical perspective, a challenge must have enough data already generated to address the scientific question. Importantly, collaborative competitions multiply the impact of the crowdsourced datasets since dozens and sometimes hundreds of researchers will look at the data to extract insights. While much data is being generated in different areas of research and business, there still remain many interesting questions for which appropriate datasets are lacking. The underlying data must be not only sufficiently large but also of sufficient quality, diversity and complexity so that researchers can extract revealing patterns from the data. Additionally, the datasets must be of sufficient size so that robust statistical evaluation can be performed.

Crowdsourcing needs crowds: the more participation in a challenge, the higher the probability of solving it. However, and perhaps perversely, if hundreds of solutions are submitted, the statistical significance of the potential solutions degrades because of multiple hypothesis testing. See to Chapter 2 for practical advice on challenge design and Chapter 3 for details on challenge preparation.

Fundamental to crowdsourced collaborative competitions is the unbiased, rigorous benchmarking of methods. This requires not only a sufficient amount and quality of data, but also a gold standard set of data to evaluate method perfor-mance. This gold standard must be inaccessible to challenge participants, which is an essential component of rigor in challenges. Also central to benchmarking is the metric used for scoring and evaluation. There are often many ways to evaluate method performance with each metric capturing a different aspect of performance. Selecting the proper metric must be considered in the context of the question being addressed. While the gold standard data and evaluation metrics establish rigor, unbiased evaluation is unique to crowdsourced challenges. In the typical research process, researchers develop methods and evaluate them on metrics and data that they select. Even unconsciously, this form of self-evaluation presents a biased method assessment, which can lead to what has been termed the "self-assessment trap." [39] Crowdsourced challenges remove this bias in favor of a proper benchmarking of methods; however, the results of a challenge must be interpreted in the context of the data, question, and scoring metric. See Chapter 4 for details on scoring metrics and challenge judging. Additionally, see Chapter 11 for challenges that have special designs, such as where teams submit an agent, instead of a discrete model, and the agent interacts with a simulator.

---

[1] https://www.kaggle.com/c/higgs-boson

## 1.5 Challenge Organization

While it may seem simple at first sight, running collaborative competitions poses significant operational problems that require a coordinated effort from the organizers. Figure 1.1 shows the typical tasks involved in a challenge which require four layers of expertise: scientific, technical, legal and ethical. The genesis of a challenge could be the existence of a dataset with complex data whose analysis could benefit from the crowdsourcing paradigm [40, 41, 42]. More often challenges arise from scientific questions for which the answer requires new method development and validation [43], or from the need to benchmark algorithms that yield divergent results and for which an objective evaluation is needed [44].

The starting point for challenge organization is the definition of the scientific question for the challenge. This question must be of an applied or basic research nature, and to be impactful its importance has to be apparent. Addressing an impactful question is important to secure funding and to motivate challenge participation. Many challenges are hosted by organizations that focus on specific types of problems, such as DREAM for biomedicine, CASP for protein structure prediction, or ChaLearn [45] for Machine Learning. Typically these organizations decide through internal vetting process what challenges are organized. Other organizations that foster the organization of challenges, such as NeurIPS, makes yearly open calls for competitions [46], and send the competition proposals for peer review to ensure that the challenge objectives are scientifically significant and of societal value. In all cases, the challenge question should be formulated in a way that it can be addressed in a collaborative competition setting, typically in the form of a predictive or classification algorithms. In some cases the question may simply be one of improving the state of the art in the performance of algorithms to solve a problem (e.g., pro-tein structure prediction). In others the question could be a novel question for which no reference exists in a specific domain. In the latter case, competition organizers may involve coordination with a steering committee of experts in the domain area (technologists, physicians, economists, researchers, etc.). Additional considerations should also be had into account if the data are from sensitive sources or the results from the challenge could be used in an unethical manner.

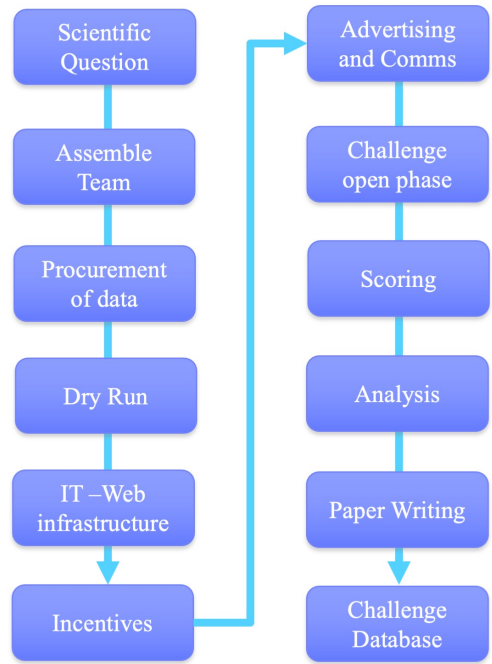

Fig. 1.1: The steps and tasks in the organization of a typical challenge.

The second step is assembling a team of data scientists, data governance and IT engineers to manage the data use agreements, data analysis tasks and IT infrastructure. It should be noted that data governance and establishing robust and reasonable data use agreements can be a time and resource consuming task, but it is essential for protection of sensitive data and to ensure its proper and ethical use.

Procurement and processing of the data is critical. It is essential that part of the data be unpublished so that it can be used as a gold standard dataset against which to score challenge submissions. This step usually requires Big Data processing such as normalization, compression, etc. If synthetic data is used for the challenge, appropriate tests are required to ensure it maintains privacy while maintaining fidelity to the original data. Having the data organized and packaged in easy-to-use datasets is necessary to reduce the barriers for participation. Adequate data governance must be in place, such as agreements with data producers, study participants, and Institutional Review Board (IRB) approvals for human data. Restrictions on the use of the data outside of the competition should be clearly stated in the competition rules.

Conducting an internal "dry-run" amongst the challenge organizing team or "mock participants" to test if the posed challenge question can be addressed with the data at hand is critical to determine if the challenge has sufficient data to solve the problem, but also that the challenge cannot be solved with trivial approaches. The dry run outcomes are (a) Dataset splits into balanced training, validation and test sets, (b) Selection of scoring metrics, (c) An estimate of the difficulty in answering the challenge questions with the data at hand (if a challenge seems impossible, then it may be better not to do it!), and (e) A definition of a baseline solution that the participants should improve upon, ideally using the present state-of-the-art approach. Further documenting the solution with background information (e.g. a challenge overview and selected literature) and step-by-step guide to submissions as part of a "starter kit" is also extremely useful in reducing the participation barrier and increasing engagement.

A challenge needs an information technology infrastructure and web content. Important ingredients of such infrastructure are a registration system (with the requirement that participants agree to the challenge terms and conditions) [47], a challenge website containing a detailed challenge description, dataset storage and download and submission uploads, leaderboards for real-time feedback of performance, and a discussion forum where participants can communicate with organizers and other participants. The DREAM Challenges, for example, use Sage Bionetworks' Synapse platform https://www.synapse.org. Finally, in our era of Big Data, challenge platforms have started to use cloud services, for example Kaggle Notebooks https://www.kaggle.com/code where models written as code notebooks are developed and can directly access both public and private data. See Chapter 5 for a review of challenge platforms.

The next step is the definition of incentives to recruit as many participants as possible to solve the challenge. Incentives can include an invitation to the best performers to help draft and submit a challenge overview paper, a speaking invitation to conferences, monetary awards, and in some challenges a job offer. Often times, individual teams are encouraged to publish their methodology as a separate publication. In the academic community the most popular incentive is the prospect of co-authoring manuscripts, but many participants are motivated to participate in a collaborative effort where they can work on interesting and unpublished data to address an impactful problem.

Before launching the challenge, it is desirable that a robust marketing and outreach campaign should be in place. This step is highly necessary as the success of the challenge depends on the degree of participation. Successful marketing approaches include the use of press releases, mailing lists, pre-challenge commentaries in top tier journals, interactive "warm up" or "kickoff" sessions, and direct outreach to researchers in the domain most directly connected to the challenge in question. See Chapter 12 for an in depth discussion on practical considerations in challenges, such as sponsors, grants, prizes, dissemination, and publicity.

After much preparation, the day arrives when the challenge is launched. The data is crowd-sourced and solutions to the scientific problems posed in the challenge are solicited in the "open phase," also referred to as the "feedback phase" or "development phase." The open phase is characterized by the improvement of algorithms and methods using leaderboards to monitor progress, an open discussion of ideas and data features, using a Discussion Forum, and a deadline to submit the solutions. To submit their solutions participants can simply send a file with their list of predictions of the gold standard target values, or executables such as docker containers that will be run on withheld data for training and/or inference. By bringing the model to the data, the latter code-submission competitions overcome the difficulties of accessing private datasets, and increase the reproducibility of the results [48]. In the open phase, limits in the number of allowed submissions are needed to prevent learning the gold standard **test data** by trial and error. The number of submissions per day or total number of submissions may be limited. Submission limits may be circumvented by coordination between teams or creating multiple registrations per participant. To avoid this, such team coordination or creating fake teams with same participants but different team names must be explicitly prohibited in the challenge rules. Depending on the organization, there are different restrictions about the copyright and intellectual property rights associated with the submissions. Organizations that foster open and collaborative science typically require that participants submit open source code and an explanation of the methods used in the predictions to be accessed publicly.

When the open phase of the challenge finishes, the final assessment phase starts, in which submissions are evaluated to determine the best performers. The best submission from the open phase or a single final submission can be used to determine the final leaderboard metrics. Scoring consists of comparing the submitted predictions of outcomes against the true outcomes using a held out gold standard, which is known to organizers but not to the participants. In order for a final score to be meaningful, it has to be accompanied with a statistical criterion of how difficult reaching that degree of performance is, typically under a null hypothesis. Several scoring metrics can be used to assess different aspects of the predictions [49].

After the final scoring phase ends, the opportunity to learn from the aggregate of submissions starts. In the DREAM Challenges, for example, organizers and interested challenge participants work together to analyze the results of the challenge and write a paper that describes it. Many challenge platforms organize a conference to discuss take home lessons, and present results. There are a few venues that focus on reporting challenges, such as the NeurIPS Competition Track and the Chalenges in Machine Learning (ChaLearn) book series http://www.chalearn.org/books.html. See Chapter 6 for tips on writing a post-challenge paper.

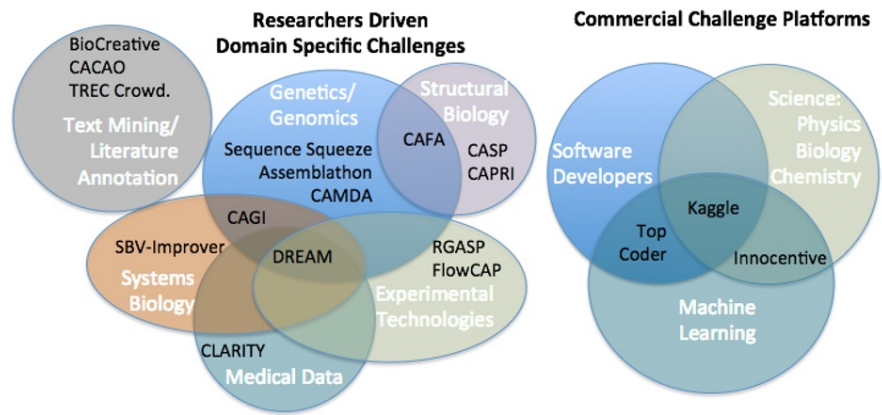

Fig. 1.2: Non-exhaustive subset of challenge platforms and organizations in the biomedicine space (left), and some platforms that organize challenges in a variety of areas of science and technology (right).

Besides papers reporting on the overall challenge, the legacy of a challenge can be curating a database that archives and makes available for future use in education, research and benchmarking the concrete outcomes of the challenge. This database typically includes software code and documentation wikis of the participants and teams who provided a final submission. Example repositories include Sage Bionetworks Synapse https://www.synapse.org, UCI machine learning repository https://archive.ics.uci.edu/ml/index.php, Kaggle datasets https://www.kaggle.com/datasets, and Papers with Code https://paperswithcode.com/

## 1.6 Challenge platforms

The success of the crowdsourcing paradigm has spurred a proliferation of challenge initiatives and platforms. Wikipedia lists close to 150 crowdsourcing projects in very diverse areas such as design and technology innovation. Figure 1.2 shows the most popular researcher driven challenge initiatives as well as the most popular commercial challenge platforms. The list is not comprehensive, but highlights consistent and well-established challenge initiatives.

Among the researcher driven challenges, biomedicine is an exemplar of using challenges to support the significant changes in research needs. With the rapid increase in the amount of data, especially molecular information on genetic sequences, signaling pathways, protein structures, and medical imaging, public challenges are a strategy to develop and share novel methods. The topics that have profited the most from these types of efforts are structural biology (CAPRI, CASP), genomics (Sequence Squeeze, Assemblathon, CAMDA), systems biology (sbv-IMPROVER), text mining (BioCreative, CACAO, TREC Crowd), medicine (CLARITY), medical imaging (MICCAI), and emerging technologies in search of benchmarking and new analytical tools (RGASP, FlowCAP). Some challenge initiatives straddle more than one domain areas such as CAFA (Genetics/Genomics, Structural Biology), CAGI (genetics/Genomics, Systems Biology) and DREAM (Genetics/Genomics, Systems Biology, Emerging Technologies and Medicine). Initiatives such as DREAM, Flow-CAP, CAGI and sbv-IMPROVER organize several challenges per year, and only the generic project and not specific challenges are shown in Figure 1.2. As mentioned earlier, the case of structural biology and CASP is a paradigmatic example of how challenges have nurtured a field. For the last three decades, scientists tackled the key problems in this field in biannual challenges, followed by conferences where results were discussed. The field steadily progressed until in 2020, a deep learning model, AlphaFold, provided spectacular results [50, 51]. While there are many open questions in the field, this was a major leap that built on the work of the community over the years. Machine learning research has similarly experienced a rapid expansion in method research, to support research in this area, the Conference on Neural Information Processing Systems (NeurIPS) has created a dedicated competition track since 2017 to highlight challenging problems and innovative solutions. See Chapter 5 for details on challenge platforms.

Crowdsourcing can be a profitable business. The business consists of organizing challenges as a fee for service for other companies that may not have the expertise necessary to give solutions to a specialized task. In such cases crowds can fill that expertise gap. Amongst the most popular and successful commercial challenge platforms we can mention Innocentive (https://www.innocentive.com/), which crowdsources challenges in science and technology (Social, Physics, Biology, Chemistry); TopCoder (https://www.topcoder.com/), which serves the software developer community; and Kaggle (https://www.kaggle.com/), that administers challenges to machine learning and computer savvy experts, addressing predictive analytics problems in a wide range of disciplines. Other open source competition platforms include Codalab [31] (https://codalab.lisn.upsaclay.fr/), EvalAI (https://eval.ai/) and Grand Challenge (https://grand-challenge.org/).

## 1.7 Conclusions/perspective

Crowdsourced challenges offer a different way of doing science or solving problems collaboratively. This is not to say better than traditional approaches, but an alternative way to engage solvers and make valuable data available to the community. As discussed earlier in this Chapter, crowdsourcing is not a new idea, though when applied to science, it does cut against the grain of traditional, siloed academic research. Ultimately, we are at a point where the generation of data is outpacing our ability to make sense of this data. Team science has grown concomitantly with the generation of these data because modern science and technology questions are often complex and require a multi-disciplinary approach. As the nature of data continues to change, both in richness and volume, analysis is a continual difficulty we face. To advance science we must explore different modes of research to expand the frontiers of knowledge.

There are opportunities for crowdsourced challenges to accelerate many aspects of academic education and research. In education, challenges in different fields can be used as learning modules to introduce computational methodologies and their rigorous evaluation. Students from high school to graduate level could develop their skills in ongoing challenges, while learning to collaborate with others to solve pressing problems in biomedicine, sustainability, environmental sciences, etc. Chapter 9 discusses the educational value of challenges and benchmarks in more detail. In research, the sheer amount of work that can be focused on one question in a short period of time is unmatched. As an illustration, a typical challenge that runs for a period of 5 months with 150 participants. Assuming that each researcher on average worked 100 hours on the challenge,represents about 15,000 hours of research effort dedicated to addressing one question; there are only 3,600 hours in 5 months. Even if an individual were to dedicate this amount of time to address a single question, it is unlikely that this individual would have the cross-disciplinary knowledge of 150 participants, thus a much smaller sampling of methods would be explored. One can imagine community efforts that both produce data and run a challenge to address a question in a timeframe shorter than even the best funded research institutions can attain. If harnessed, this energy could potentially impart an extraordinary increase in the velocity and depth with which important problems are attacked. Subsequent chapters will provide overviews and special considerations for academic, industrial and educational challenges.

Crowdsourced challenges produce rigorous, unbiased benchmarked data and methods, and results represent the state-of-the-art in the field. Best performing methods produced in any given challenge have undergone a vetting process that cannot be done by any individual research group. This vetting process can be used to aid in academic publications, or a challenge-assisted peer review. Successfully applied by the journal Science Translational Medicine in the Sage-BCC DREAM Challenge [52], this form of peer review is more rigorous than any individual reviewer can do.

Data is being generated at an unprecedented rate and the number of potential crowdsourced challenges is increasing. This presents a challenge in and of itself. The selection of challenges will become increasingly difficult with many worthy challenges competing for the attention of the research community and the research community growing more fatigued of these challenges. It remains an open question how to most effectively select questions for the community to address, with a potential solution being that the community itself crowdsources the question, that is, let the community decide which challenge to address. Without participation, crowdsourced challenges would not exist, so for that, we thank all the curious and ambi-tious solvers that have contributed to these community efforts. See Chapter 10 to get started on creating your own challenge or benchmark.